\documentclass{article}

\PassOptionsToPackage{utf8}{inputenc}
\usepackage[utf8]{inputenc}
\newcommand{\setk}[1]{\{0,1,\ldots,#1\}}
\usepackage{amsmath,amsfonts,bm}

\def\eqref#1{equation~\ref{#1}}

\def\1{\bm{1}}

\def\mA{{\bm{A}}}

\def\mW{{\bm{W}}}
\def\mX{{\bm{X}}}

\DeclareMathAlphabet{\mathsfit}{\encodingdefault}{\sfdefault}{m}{sl}
\SetMathAlphabet{\mathsfit}{bold}{\encodingdefault}{\sfdefault}{bx}{n}

\def\gN{{\mathcal{N}}}

\usepackage[export]{adjustbox}
\usepackage{multirow}
\usepackage{amsmath}
\usepackage{longtable}
\usepackage{advdate}

\usepackage{mathtools}
\usepackage{subcaption}
\usepackage{numprint}
\newcommand{\comment}[1]{}
\usepackage{amsthm}
\usepackage{float}
\usepackage{placeins}
 \usepackage{subcaption}
\usepackage{graphicx}

\usepackage{hyperref}
\usepackage{todonotes}
\usepackage[utf8]{inputenc} 
\usepackage[T1]{fontenc}    
\usepackage{hyperref}       
\usepackage{url}            
\usepackage{booktabs}       
\usepackage{amsfonts}       
\usepackage{nicefrac}       
\usepackage{microtype}      
\usepackage{xcolor}         
\newcommand{\layersuperscript}{l}
\usepackage[export]{adjustbox}
\usepackage{multirow}
\usepackage{amsmath}
\usepackage{advdate}
\usepackage{mathtools}
\usepackage{subcaption}
\usepackage{placeins}
\usepackage{blindtext}
\usepackage{amsthm}
\usepackage[scientific-notation=true]{siunitx}

\usepackage{multicol}
\usepackage{multirow}
\newcommand{\neighborhood}[1]{\gN({#1})}

\usepackage{xcolor}
\usepackage{soul}
\usepackage{url}
\usepackage{inputenc}
\usepackage{amsmath}
\usepackage{amsthm}
\usepackage{booktabs}
\usepackage{algorithm}
\usepackage{algorithmic}
\usepackage{booktabs}
\usepackage{nccmath}
\newcommand\numberthis{\addtocounter{equation}{1}\tag{\theequation}}
 \usetikzlibrary{positioning,shadows,backgrounds}

\usepackage{arxiv}
\usepackage[title]{appendix}

\usepackage[utf8]{inputenc} 
\usepackage[T1]{fontenc}    
\usepackage{hyperref}       
\usepackage{url}            
\usepackage{booktabs}       
\usepackage{amsfonts}       
\usepackage{nicefrac}       
\usepackage{microtype}      
\usepackage{lipsum}		
\usepackage{graphicx}
\usepackage[numbers]{natbib}
\usepackage{doi}

\title{Explainable Multilayer Graph Neural Network for Cancer Gene Prediction}

\author{{Michail Chatzianastasis} \\
	LIX, École Polytechnique, IP Paris\\
	Palaiseau, France \\
	\texttt{michail.chatzianastasis@polytechnique.edu} \\
	\And
	{Michalis Vazirgiannis} \\
	LIX, École Polytechnique, IP Paris\\
	Palaiseau, France\\
	\texttt{mvazirg@lix.polytechnique.fr} \\
        \And 
        {Zijun Zhang} \\
        Division of Artificial Intelligence in Medicine \\
        Cedars-Sinai Medical Center, Los Angeles, CA  \\
        \texttt{zijun.zhang@cshs.org} \\
}

\date{}

\renewcommand{\headeright}{}
\renewcommand{\undertitle}{}
\renewcommand{\shorttitle}{}

\begin{document}
\maketitle

\begin{abstract}
The identification of cancer genes is a critical yet challenging problem in cancer genomics research. 
Existing computational methods, including deep graph neural networks, fail to exploit the multilayered gene-gene interactions or provide limited explanation for their predictions.
These methods are restricted to a single biological network, which cannot capture the full complexity of tumorigenesis. Models trained on different biological networks often yield different and even opposite cancer gene predictions, hindering their trustworthy adaptation.
Here, we introduce an Explainable Multilayer Graph Neural Network (EMGNN) approach to identify cancer genes by leveraging multiple gene-gene interaction networks and pan-cancer multi-omics data.
Unlike conventional graph learning on a single biological network, EMGNN uses a multilayered graph neural network to learn from multiple biological networks for accurate cancer gene prediction. 
Our method consistently outperforms all existing methods, with an average 7.15\% improvement in area under the precision-recall curve (AUPR) over the current state-of-the-art method.
Importantly, EMGNN integrated multiple graphs to prioritize newly predicted cancer genes with conflicting predictions from single biological networks. For each prediction, EMGNN provided valuable biological insights via both model-level feature importance explanations and molecular-level gene set enrichment analysis. 
Overall, EMGNN offers a powerful new paradigm of graph learning through modeling the multilayered topological gene relationships and provides a valuable tool for cancer genomics research.
\end{abstract}

\section{Introduction}

Understanding the precise function and disease pathogenicity of a gene is dependent on the target gene's properties, as well as its interaction partners in a disease-specific context \citep{sealfon2021machine,zitnik2019machine,greene2015understanding}. 
High-throughput experiments, such as whole-genome sequencing and RNA sequencing of bulk and single-cell assays, have enabled unbiased profiling of genetic and molecular properties for all genes across the genome. 
Experimental methods to probe both physical \citep{qin2021deciphering,bruckner2009yeast}  and genetic interactions \citep{norman2019exploring,costanzo2010genetic} provide valuable insights of the functional relevance between a pair of genes. 
Based on these data, computational methods have been developed to predict gene functions for understudied and uncharacterized genes by combining the gene's property with its network connectivity patterns \citep{berardini2004functional,ietswaart2021genewalk,pfeifer2022gnn}. 
However, the prediction of gene pathogenicity in disease-specific contexts is challenging. 
Functional assays describing the gene and its gene network are relevant to disease only to the degree to which the measured property correlates with disease physiology \citep{nykamp2017sherloc}; while our understanding of complex disease physiology is poor, 
even for diseases with large sample size and data modalities, such as cancer \citep{liu2018integrated}.

As the completeness of known cancer genes is questioned, predicting novel cancer genes remains a crucial task in cancer genomics research. 
These genes, which are often mutated or aberrant expressed in cancer cells, play a key role in the development and progression of the disease \citep{sondka2018cosmic}.
Large-scale cancer sequencing consortia projects have generated genomic and molecular profiling data for a variety of cancer types, providing an information-rich resource for identifying novel cancer genes.
Building on the hypothesis that pan-cancer multi-omic modalities provide critical information to cancer gene pathogenicity, a pioneering work EMOGI~\citep{schulte2021integration} innovatively modeled the multi-omics features of cancer genes in Protein-Protein interaction (PPI) networks to predict novel cancer genes. To address the challenge of functional properties irrelevant to cancer disease physiology, EMOGI featurized each gene by a vector summarizing multi-omics data levels across various cancer types in The Cancer Genome Atlas (TCGA)~\citep{weinstein2013cancer}. EMOGI then modeled the gene-gene interactions from pre-defined generic PPI networks using a Graph Convolution Neural network (GCN). When trained on a set of high-confidence cancer- and non-cancer genes, EMOGI identified $165$ newly predicted cancer genes without necessarily recurrent alterations, but interact with known cancer genes. 

A major limitation of EMOGI is that it didn't address the disease physiology relevance in the pre-defined graph topology and connectivity patterns.
EMOGI employed six different pre-defined graphs, including genetic-focused networks such as Multinet~\citep{khurana2013interpretation}, and generic protein interaction networks such as STRING-db~\citep{szklarczyk2019string}. 
Among EMOGI models trained on different PPI networks, we found an average standard deviation of $25.2\%$ in unlabelled cancer gene predictions, demonstrating the newly predicted cancer genes were different when using different PPI networks.
Thus, a trustworthy adaptation of the EMOGI method's output is challenging because conflicting prediction results are ubiquitous. 
Furthermore, as cancer disease physiology is complex, using a single predefined graph to represent the gene-gene relationships cannot fully capture its molecular landscape; therefore, more sophisticated, data-driven methods are needed to decipher the gene relationships in disease-specific contexts.

Here, we propose a novel graph learning framework, EMGNN (Explainable Multilayer Graph Neural Network), for predicting gene pathogenicity based on multiple input biological graphs. 
EMGNN maximizes the concordance of functional gene relationships with the unknown disease physiology by jointly modeling the multilayered graph structure.
We evaluated the performance of EMGNN in predicting cancer genes using the same compiled datasets as EMOGI and showed that our proposed method achieves state-of-the-art performance by combining information from all six PPI networks.
Furthermore, we explained EMGNN's prediction by both model-level integrated gradients and molecular-level gene pathways. 
By examining newly predicted cancer genes identified by EMGNN, we demonstrated biological insights can be revealed by leveraging the complementary information in different types of biological networks.
Overall, EMGNN provides a powerful new paradigm of graph learning through modeling the multilayered topological gene relationships. Our key contributions can be summarized as follows: 
\begin{itemize}
\item We develop an Explainable Multilayer Graph Neural Network (EMGNN) approach to identify cancer genes by leveraging multiple protein-protein interaction networks and multi-omics data.
\item Our method demonstrates superior performance compared to existing approaches as quantified by a significant increase in the AUPRC across six PPI networks. The average improvement in performance is 7.15\% over the current state-of-the-art method, EMOGI.
\item We identify the most important multi-omics features for the prediction of each cancer gene, as well as the most influential PPI networks, using model interpretation strategies. 
\item EMGNN identifies newly predicted cancer genes by integrating multiple PPI networks, providing a unified and robust prediction for novel cancer genes discovery. Our code is publicly available on GitHub\footnote{Code: \url{https://github.com/zhanglab-aim/EMGNN}}
\end{itemize}

\begin{figure*}[t]
    \centering
    \includegraphics[width=\textwidth]{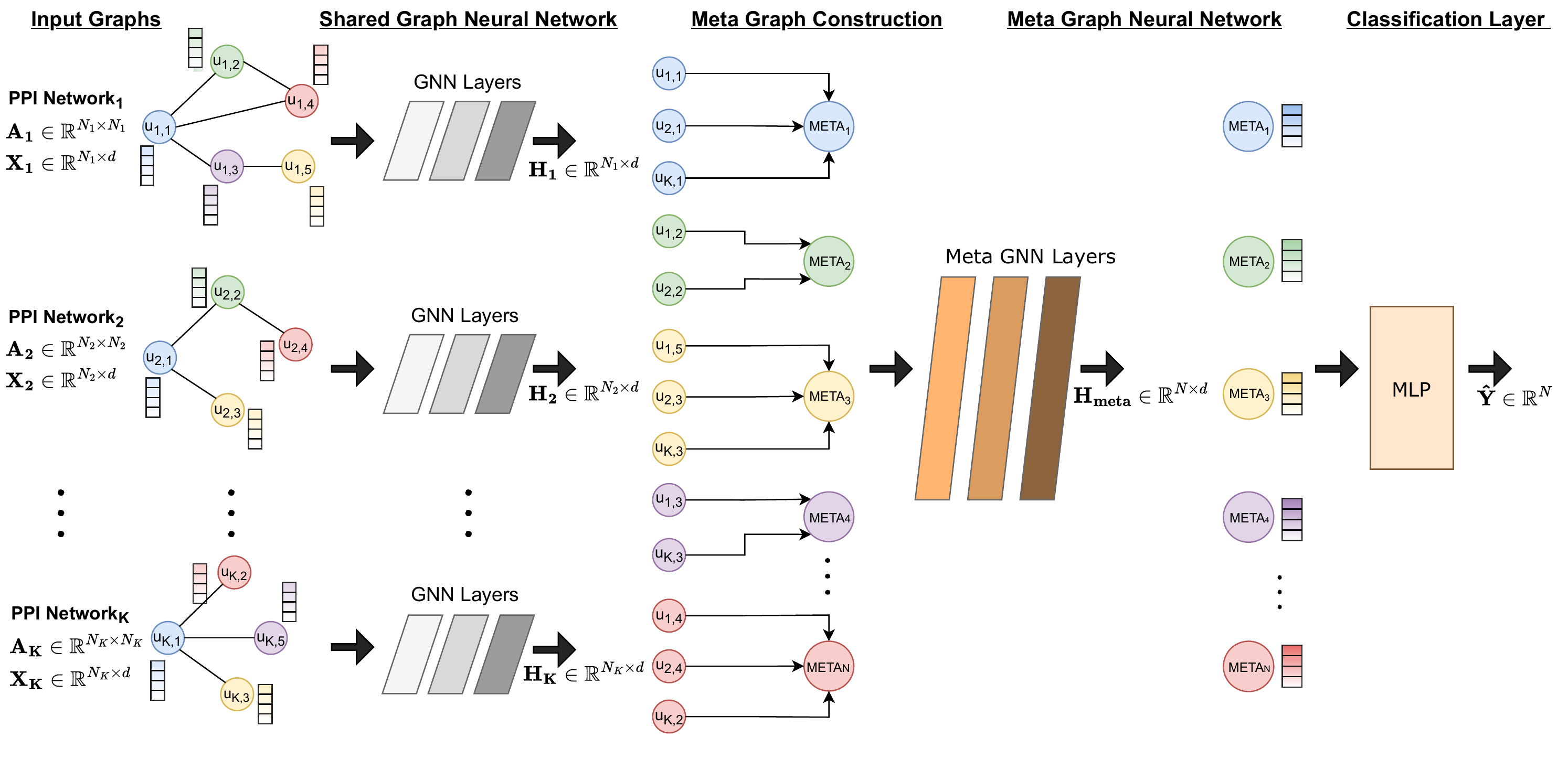}
    \caption{An illustration of our proposed Explainable Multilayer Graph Neural Network (EMGNN) approach. The model consists of three main steps: (1) Apply a shared GNN to update the node representation matrix of each input graph; (2) Construct a meta graph for each gene, where the same genes across all graphs are connected to a meta node, and update the representation of the meta nodes with a second GNN, Meta GNN; (3) Use a multi-layer perceptron to predict the class of each meta node.}
    \label{fig1}
\end{figure*}

\section{Results}
\subsection{Overview of EMGNN model}

To effectively model multilayered graph structures for cancer gene prediction, we developed a graph neural network (GNN) model EMGNN that jointly learned from multiple biological networks as inputs. 
The extension of GNNs to handle multiple networks is a non-trivial task, as they are designed to operate on a single graph. 
The input for EMGNN is multiple graphs where each graph describes gene-gene relationships by an adjacency matrix, and a feature matrix for genes that is shared by multiple graphs.
The output is a binary node classification of putative cancer vs non-cancer gene. 
EMGNN achieves multilayered graph modeling in a three-step approach (Figure \ref{fig1}).
In the first step, EMGNN updates the node representation within each graph layer (i.e., a single biological network) by a shared message passing operator. 
As different layer-wise graphs have distinct connectivity patterns, node representations will be updated differently in each layer.
The shared message passing operator allows the EMGNN model to incorporate new graphs as inputs, while keeping the model's trainable parameters fixed.

EMGNN then integrates the layer-wise node representations of the same genes across multiple biological networks by a meta graph (Methods). For each gene, the meta-graph consists of a meta-node for this gene, which receives messages across individual networks from the given gene's representations. This message passing step is modeled with a second GNN, referred to as the Meta GNN (Figure \ref{fig1}). 
Meta GNN enables directed message-passing to combine and exchange information from different networks to the meta nodes, which will contain the final representations of the genes. 
In the last step, a multi-layer perceptron (MLP) takes the meta node representations as input, and performs the node classification task. 

Notably, our EMGNN model is a generalized, multilayered form of a single graph GNN. 
In the special cases where multilayered graphs have identical adjacency matrices, or only one graph is provided as inputs, EMGNN reduces to a standard single-graph GNN, where the shared message passing operators are standard GNN operators, and meta GNN reduces to an identity operation. 
Thus, EMGNN generalizes single-graph GNN by jointly learning from the complementary information stored in multiple graphs.

\begin{table}[t]
\centering
    \caption{Test AUPRC values and standard deviation across different PPI networks across five different runs.}
    \label{table1}
    \resizebox{\columnwidth}{!}{
    \begin{tabular}{lcccccc}
    \toprule
    \textbf{Method} & \textbf{CPDB}  & \textbf{Multinet} & \textbf{PCNet} & \textbf{STRING-db} & \textbf{Iref} &
    \textbf{Iref(2015)} \\
    \midrule
    Random & 0.27 & 0.18 & 0.14 & 0.24 & 0.17 & 0.28 \\
    20/20+ & 0.66 & 0.62 & 0.55 & 0.67 & 0.61 & 0.65 \\
    MutSigCV & 0.38 & 0.33 & 0.27 & 0.41 & 0.35 & 0.43\\ 
    HotNet2 diffusion & 0.62 & 0.56 & 0.48 & 0.50 & 0.45 & 0.65 \\
    DeepWalk+features RF & 0.74 & 0.71 & 0.72 & 0.71 & 0.66 & 0.71 \\
    PageRank & 0.59 & 0.53 & 0.54 & 0.44 & 0.42 & 0.62 \\
    GCN without omics & 0.57 & 0.53 & 0.47 & 0.39 & 0.37 & 0.64 \\
    DeepWalk + SVM & 0.73 & 0.51 & 0.63 & 0.52 & 0.62 & 0.66 \\
    RF & 0.60 & 0.59 & 0.51 & 0.61 & 0.54 & 0.62 \\
    MLP & 0.58 & 0.63 & 0.47 & 0.63 & 0.55 & 0.64 \\
    EMOGI\cite{schulte2021integration} & 0.74 & 0.74 & 0.68 & 0.76 & 0.67 & 0.75\\
    EMOGI\cite{hong2022reusability} & 0.775$\pm$0.003  & 0.732$\pm$0.003 & 0.745$\pm$0.002 & 0.763$\pm$0.003 & 0.701$\pm$0.004 & 0.757$\pm$0.001 \\
    \midrule
    \textbf{EMGNN(GCN)} & \textbf{0.809} $\pm$ 0.006 & \textbf{0.854} $\pm$ 0.007 & \textbf{0.761} $\pm$ 0.001 & \textbf{0.856} $\pm$ 0.002 & \textbf{0.822} $\pm$ 0.002 & \textbf{0.800} $\pm$ 0.010   \\
    \textbf{EMGNN(GAT)} & 0.776 $\pm$ 0.018 & 0.796 $\pm$ 0.034 & 0.730 $\pm$ 0.031 & 0.805 $\pm$ 0.030 & 0.739 $\pm$ 0.033 & 0.773 $\pm$  0.049\\
    \bottomrule
    \end{tabular}
    }
\end{table}

\begin{table}[t]
\centering
    \caption{Test AUPRC and standard deviation of EMGNN(GCN) for different input perturbations methods across three different runs.}
    \label{table2}
    \resizebox{\columnwidth}{!}{
    \begin{tabular}{lcccccc}
    \toprule
    \textbf{Method} & \textbf{CPDB}  & \textbf{Multinet} & \textbf{PCNet} & \textbf{STRING-db} & \textbf{Iref} &
    \textbf{Iref(2015)} \\
    \midrule
    \textbf{Random Features} & 0.703  $\pm$ 0.001  & 0.727  $\pm$ 0.002 & 0.615 $\pm$ 0.009 & 0.745 $\pm$ 0.002 & 0.674 $\pm$ 0.001 & 0.697 $\pm$ 0.005 \\ 
    \textbf{All-one Features} & 0.726 $\pm$ 0.002 & 0.769 $\pm$ 0.001 & 0.657 $\pm$ 0.010 & 0.779 $\pm$ 0.010 & 0.710 $\pm$ 0.015 & 0.725 $\pm$ 0.013 \\ 
    \midrule
    \textbf{Edge Removal(0.2)} & 0.800 $\pm$ 0.007 & 0.841 $\pm$ 0.016 & 0.746 $\pm$ 0.017 & 0.841 $\pm$ 0.009 & 0.796 $\pm$ 0.005 & 0.786 $\pm$ 0.011 \\
    \textbf{Edge Removal(0.4)} & 0.795 $\pm$ 0.004 & 0.834 $\pm$ 0.009 & 0.743 $\pm$ 0.003 & 0.828 $\pm$ 0.004 & 0.790 $\pm$ 0.012 & 0.802 $\pm$ 0.006\\
    \bottomrule
    \end{tabular}
    }
\end{table}

\subsection{Multilayered graph improves EMGNN performance}

We applied EMGNN to predict cancer genes due to the wealth of multi-omic profiling data available for cancer, yet the underlying tumorigenesis is highly complex and cannot be fully captured by a single biological network.
To ensure fair comparisons, we used a compiled dataset and kept the identical training/test split from a previous report \citep{schulte2021integration} (Methods).
In total, this dataset consisted of 887 labeled cancer genes, 7753 non-cancer genes and 14019 unlabeled genes, along with six PPI networks with multi-omics features. The detailed numbers of labeled positive and negative genes for training the EMGNN model in each PPI network can be found in Supplementary Table 1. 

The integration of multiple graphs leads to substantial improvements in the performance of EMGNN. 
We trained EMGNN models and evaluated the testing performance with respect to different numbers of PPI networks.
While we added other networks to the training and validation set, we held-out the same testing set from the combined training set and kept the test set identical to previous works \citep{hong2022reusability, schulte2021integration}.
An illustration of the process of adding a new biological network and a definition of training and validation splits are shown in Supplementary Figure 1. 
As shown in Figure \ref{fig2}, the testing performance increased for each of the six testing datasets as the number of input networks increased.
For Multinet, STRING-db, IRef, IRef(2015) testsets, the incorporation of more graphs steadily increased the performance without reaching a plateau.
For PCNet testset, EMGNN achieved the best testing performance by combining four PPI networks.
Because PCNet is the most densely connected network, this behavior is consistent with previously reported benchmarking results, which suggests that the performance scale with the network size \citep{huang2018systematic}.

EMGNN trained by incorporating all six graphs achieved state-of-the-art performance for all six test sets (Table \ref{table1}).
As each test set was an independent set of held-out labeled cancer and non-cancer genes from each network and we kept the set identical to previous reports (Methods), the testing performance will inform generalization error of the trained EMGNNs and provide fair comparisons to previous results.
EMGNN outperformed the current state-of-the-art method EMOGI by an average margin of 7.15\% AUPRC across all test sets, with the largest gain of 11.1\% in performance observed in the old version Iref.
The smallest gain was observed in PCNet, likely because PCNet is already an expert-assembled graph combining the information from the other five graphs \citep{huang2018systematic}. 
Nevertheless, for PCNet test set, EMGNN combining six graphs is significantly more accurate than EMOGI using PCNet (p-value=0.012, t-test).
Therefore, incorporating information from multiple networks leads to enhanced predictive power for gene pathogenicity prediction.

\begin{figure*}[t]
    \centering
    \begin{subfigure}{0.32\textwidth}
    \includegraphics[width=\textwidth]{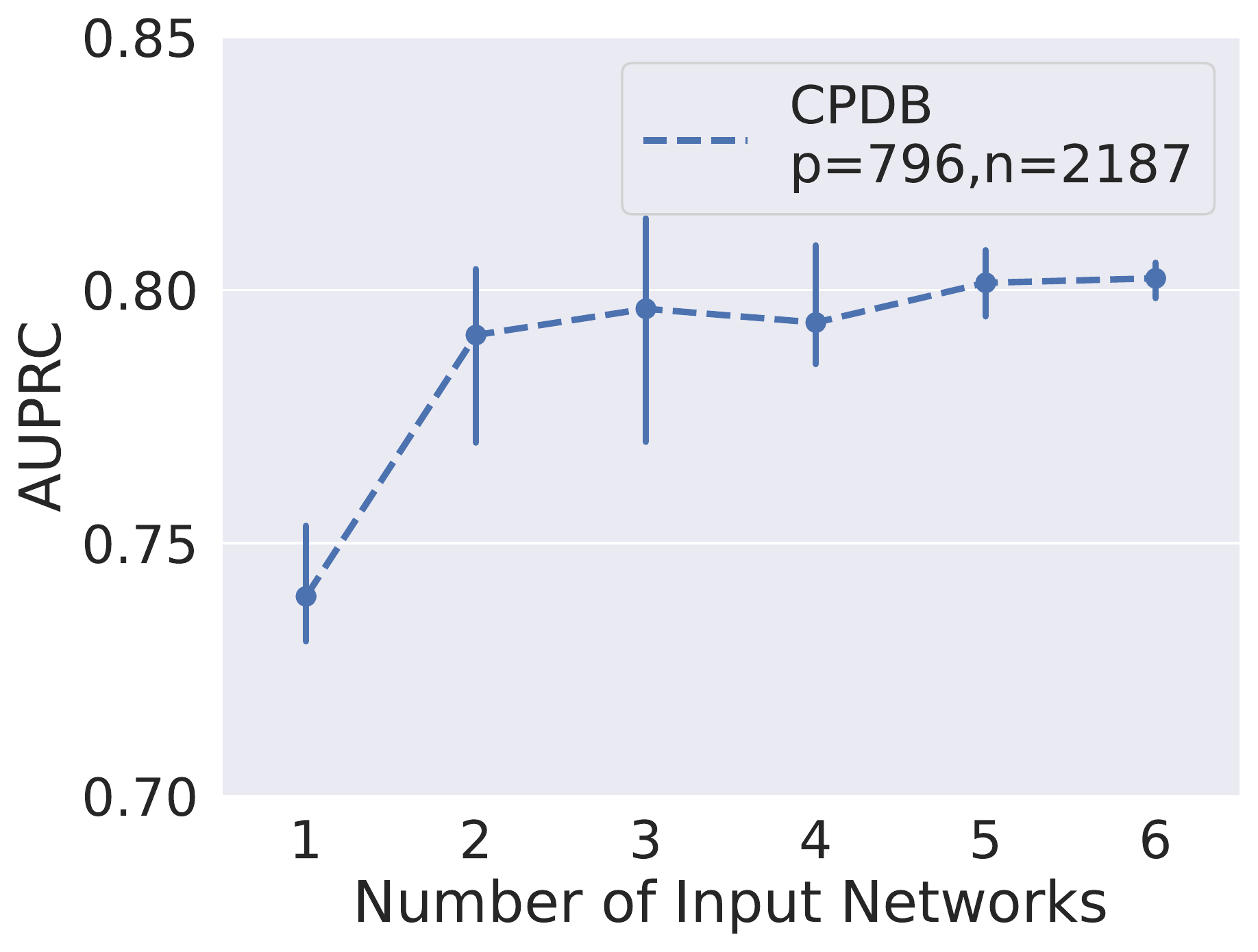}
    \end{subfigure}
    \begin{subfigure}{0.32\textwidth}
    \includegraphics[width=\textwidth]{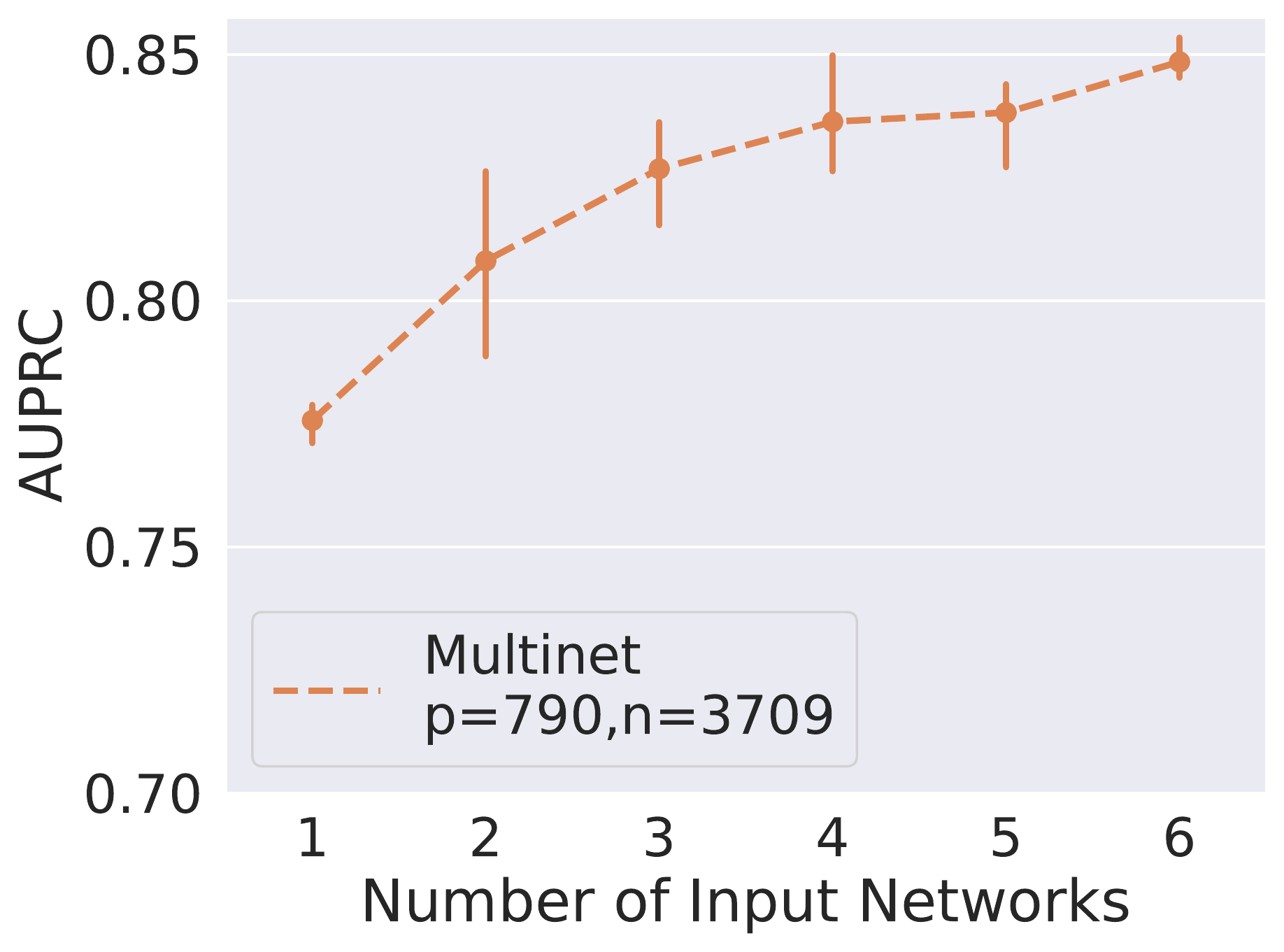}
    \end{subfigure}
    \begin{subfigure}{0.32\textwidth}
    \includegraphics[width=\textwidth]{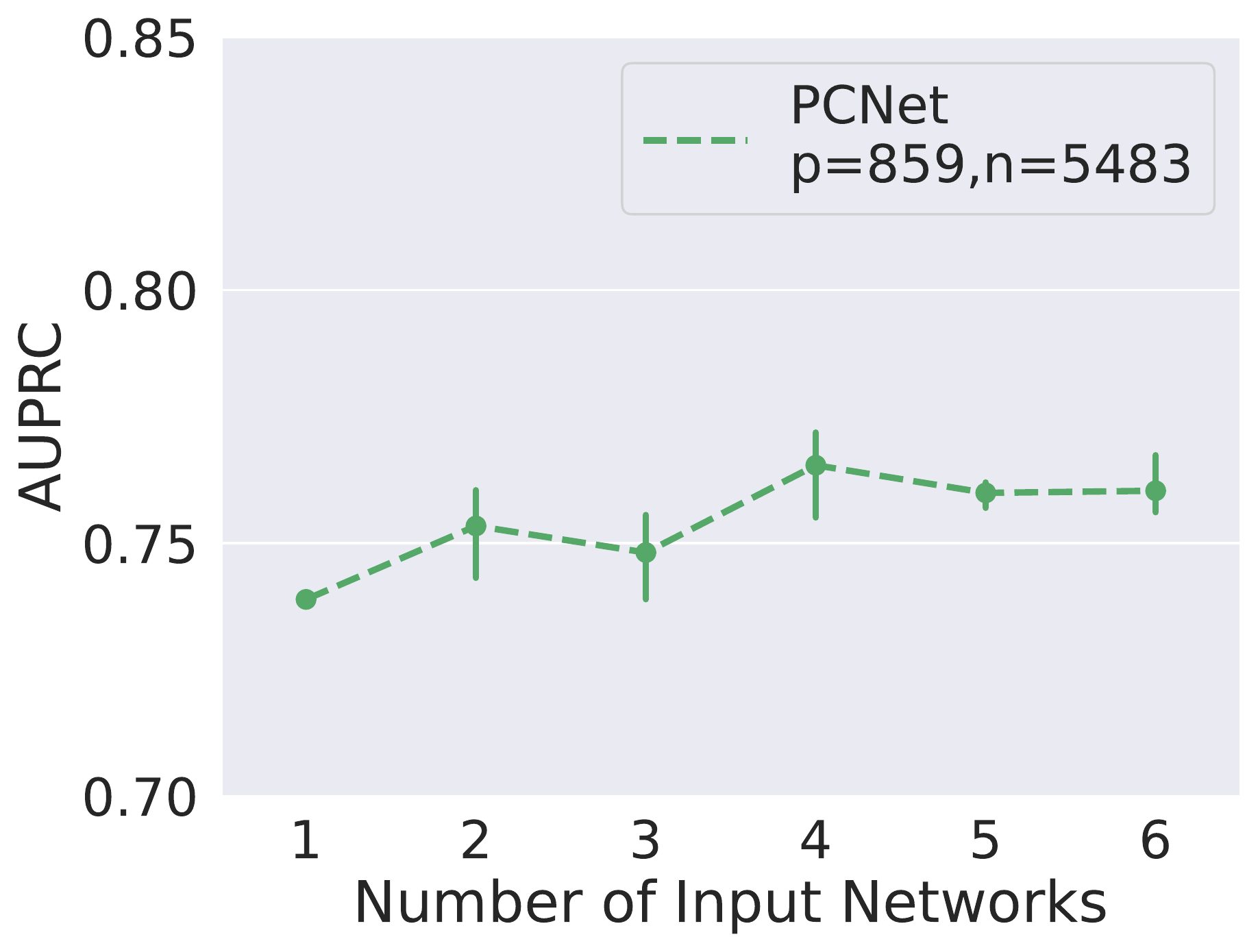}
    \end{subfigure}
    \begin{subfigure}{0.32\textwidth}
    \includegraphics[width=\textwidth]{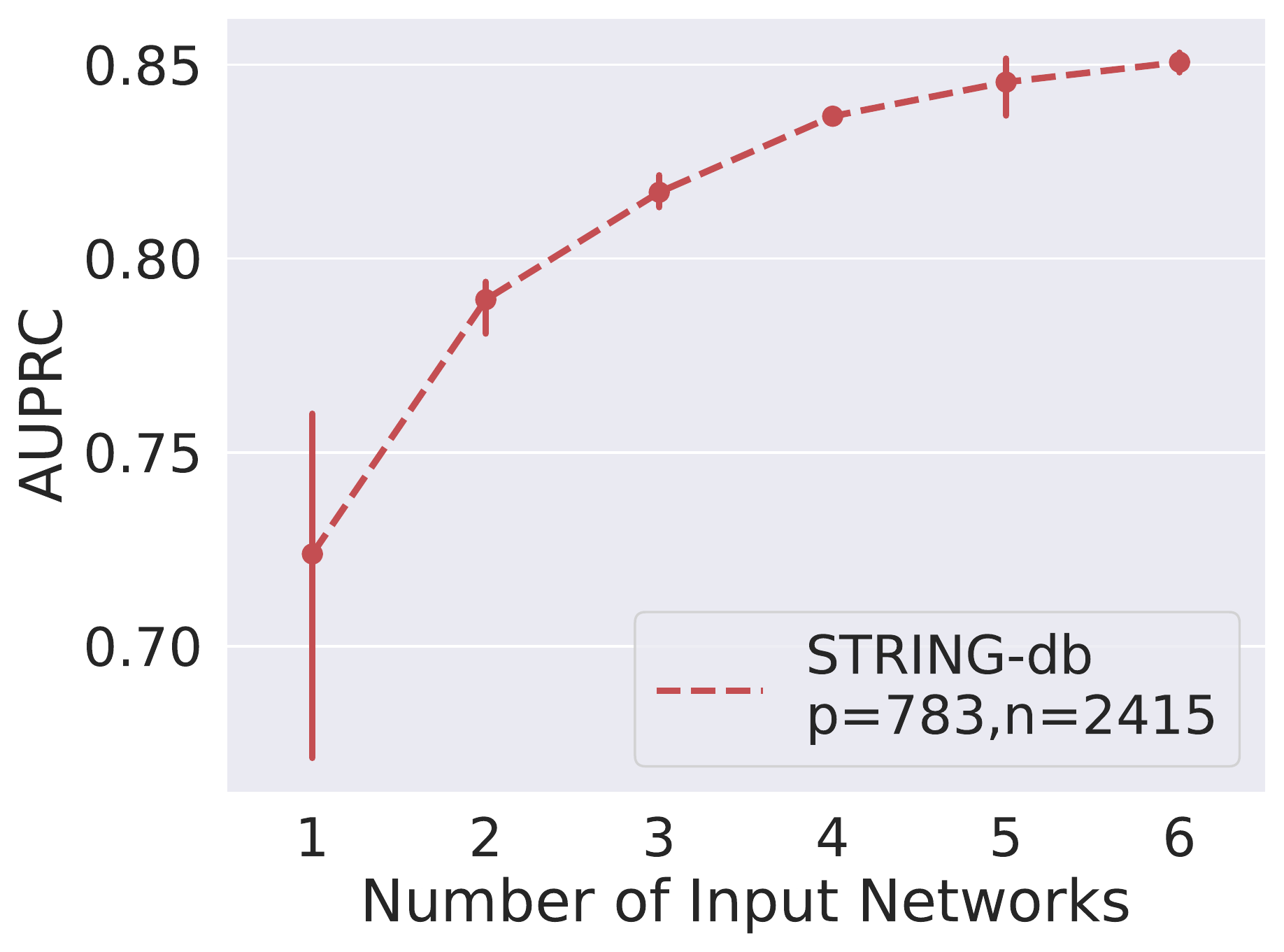}
    \end{subfigure}
    \begin{subfigure}{0.32\textwidth}
    \includegraphics[width=\textwidth]{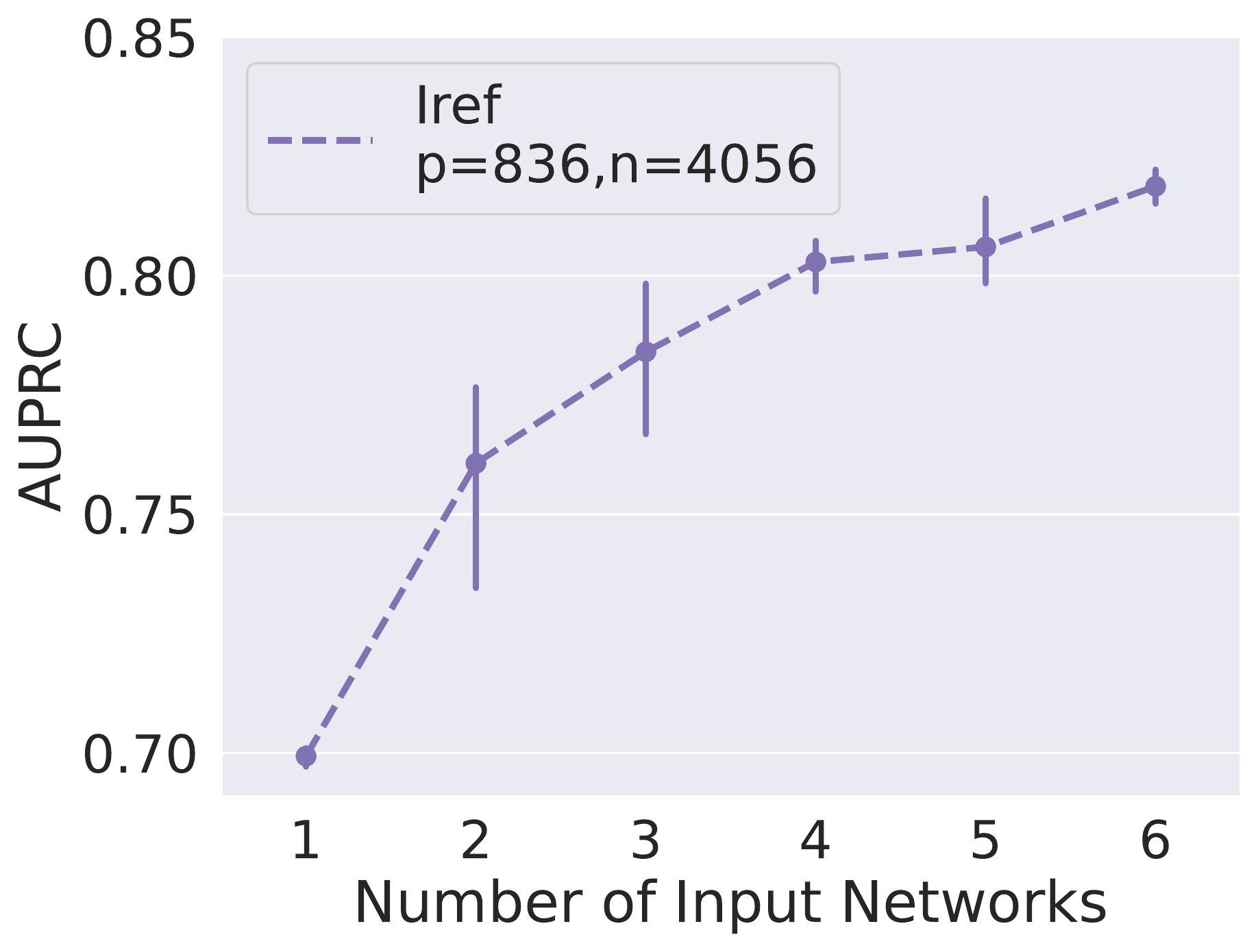}
    \end{subfigure}
    \begin{subfigure}{0.32\textwidth}
    \includegraphics[width=\textwidth]{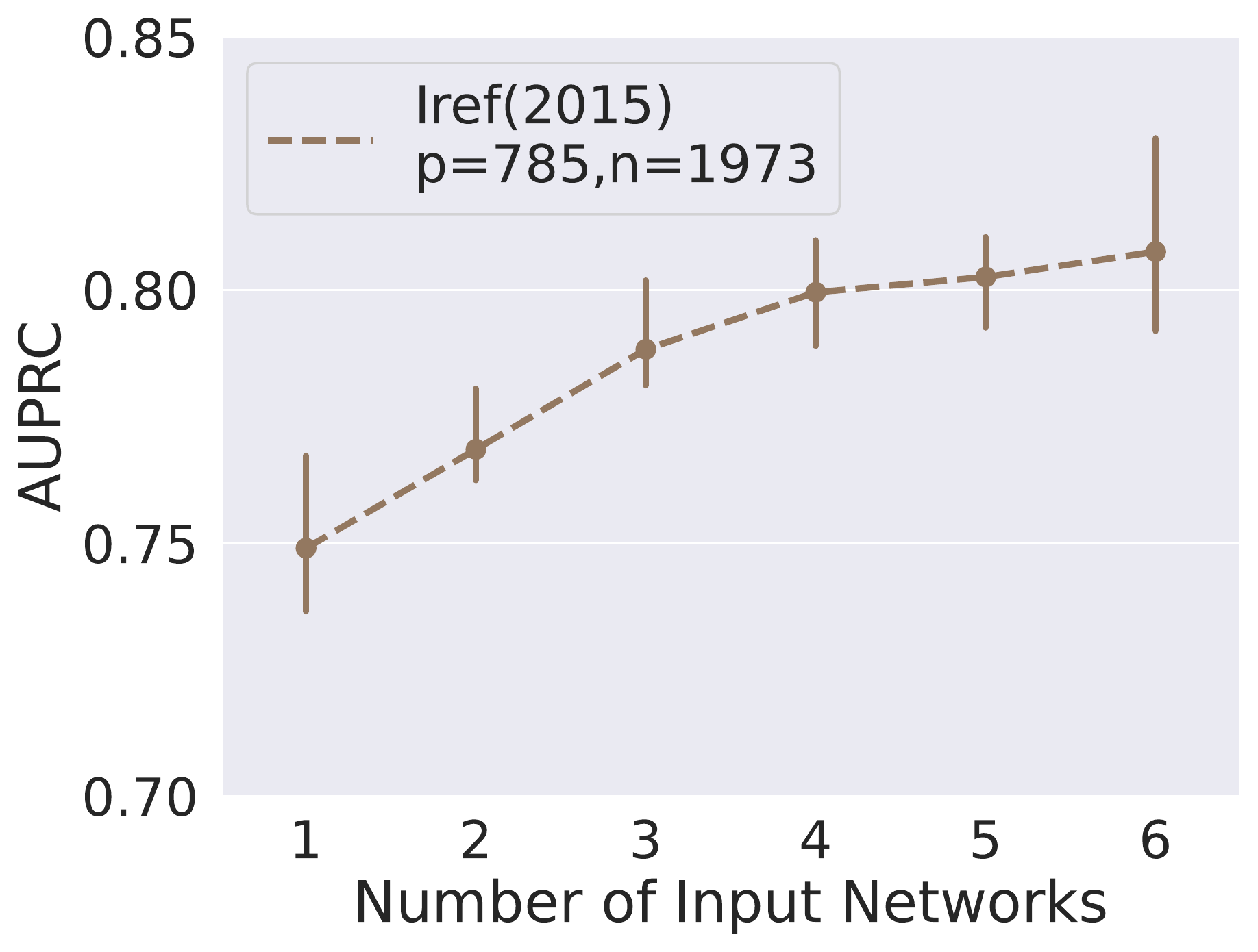}
    \end{subfigure}
    \caption{Test AUPRC and standard deviation values of EMGNN(GCN) with respect to the number of input PPI networks. Each line represents a test set of positive and negative labeled genes held out in a specific PPI network. The addition of PPI networks was conducted using a random sampling approach, where three combinations of PPI networks were sampled randomly at each point.
    Note that the testing nodes remain the same as more networks are added.
    We observe that the performance increased for the majority of the test datasets, as the number of input networks increases.}
    \label{fig2}
\end{figure*}

\begin{figure*}[t]
    \centering
    \includegraphics[width=\textwidth]{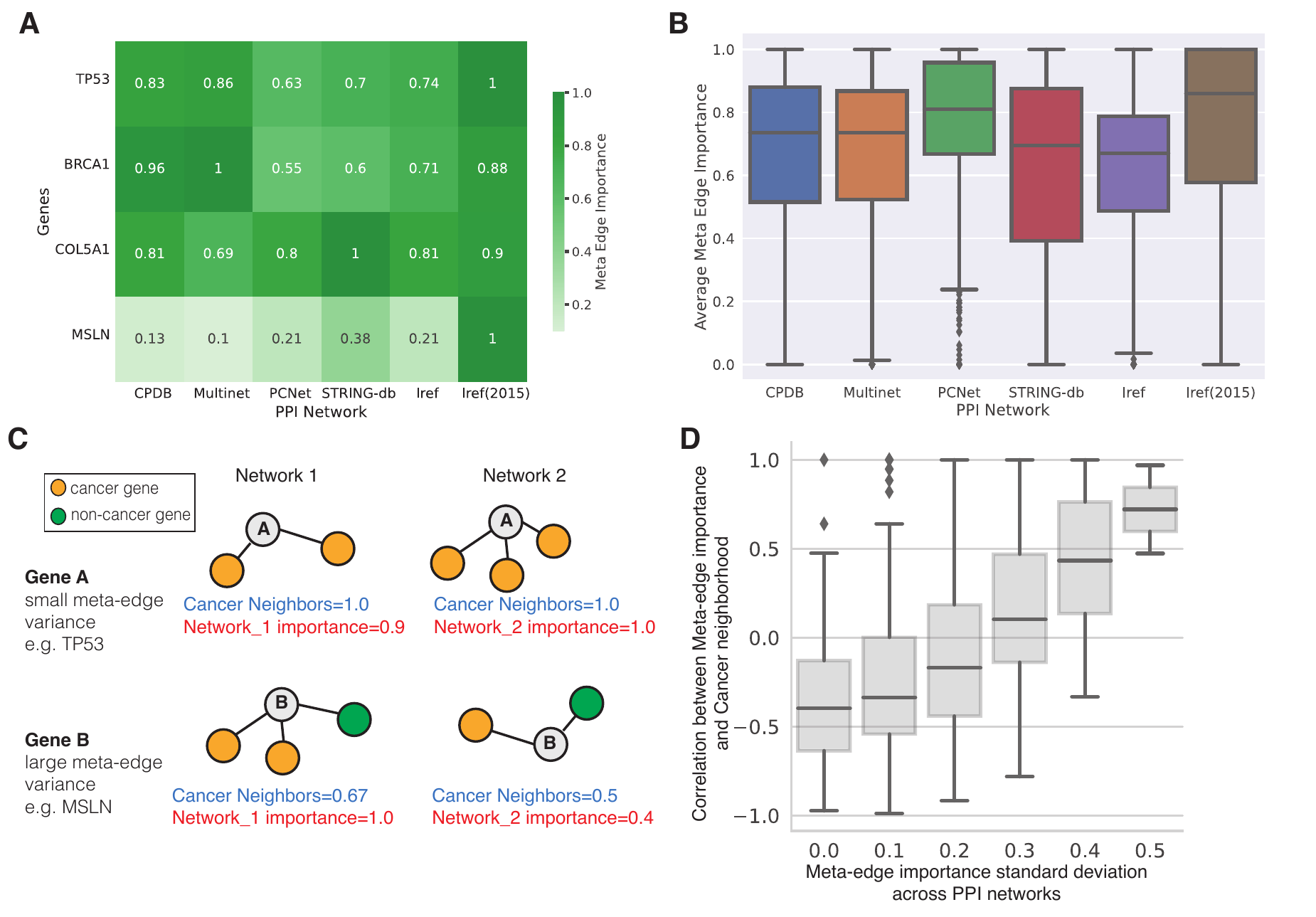}
    \caption{Explanation of each PPI network's contribution to cancer gene predictions. 
    A) Representative PPI network contributions in known cancer genes and newly predicted cancer genes. TP53 and BRCA1 are known cancer genes; COL5A1 and MSLN are newly predicted cancer genes. 
    B) Overall distribution of meta-edge feature importance for all known cancer genes across six PPI networks. Meta-edge feature importance was normalized to 1 (see Methods for details).
    C) An hypothetical illustration of PPI network cancer neighborhood implicated in the variation of meta-edge importance. 
    D) Empirical analysis demonstrates a higher correlation between meta-edge importance and cancer neighborhood for genes with a large meta-edge variance.
    }
    \label{fig3}
\end{figure*}






\subsection{Evaluating the performance of different GNN architectures and graph ablations.}

To understand each model component’s contribution to EMGNN's superior performance and model robustness, we next performed an ablation study using different GNN architectures and input perturbations.
For GNN architectures, we compared Graph Convolutional Network (GCN) \citep{kipf2017semisupervised}, 
and Graph Attention Network (GAT)\citep{velikovi2017graph}. 
After identifying the most effective GNN architecture, we proceeded to evaluate its performance under varied input conditions, including the incorporation of random and constant features, and the random removal of 20\% and 40\% of edges within the input graphs.

As shown in Table \ref{table1}, the GNN architecture played an essential role in EMGNN testing performance.
We observed that GCN is the best-performing GNN architecture in all test datasets.
Our findings demonstrated that the choice of GNN architecture has a significant impact on the performance of our model. 
Thus, EMGNN refers to EMGNN(GCN) throughout the paper unless otherwise stated.

Biologically, the EMGNN node features and edges are determined using high-throughput assays that inherently have measurement errors. To this end, we sought to answer the two following questions: How robust is EMGNN to node feature and graph structure perturbations? Are the node features and the graph structure both crucial for the prediction of cancer genes? We examined the performance of EMGNN under different types of input perturbations (Table \ref{table2}), including removal of the multi-omics node features and edges, and adding uninformative vectors (random or constant). 

We found that EMGNN decreased in performance for both random and all-one node features, suggesting that the node features derived from TCGA consortia were informative and highly relevant to cancer pathophysiology and cancer gene pathogenicity.
For edge ablations, we randomly removed 20\% and 40\% edges in each PPI network. 
The removal of edges slightly decreased EMGNN performance. 
This is expected, because EMGNN integrates six PPI networks and demonstrates its robustness towards connectivity pattern by jointly modeling a multilayered graph topology.
Overall, EMGNN effectively leveraged both node features and edges to achieve accurate predictions.

\subsection{Explaining EMGNN reveals biological insights of cancer gene pathogenicity}

Explainable and trustworthy models are essential for understanding the biological mechanisms of known cancer genes and facilitating the discovery of novel cancer genes. Given EMGNN's accurate and robust predictive performance, we developed Integrated Gradients methods~\citep{kokhlikyan2020captum} to explain the node and edge attributions of EMGNN (Methods).

A unique advantage of EMGNN is its integration of multiple PPI networks; therefore, we focused our analysis on the relative contributions from each PPI network to the known cancer gene predictions (Figure \ref{fig3}A).
Each PPI network's importance was measured by the corresponding meta edge importance (Methods).
We examined the relative contributions for two well-known cancer genes (TP53 and BRCA1, predicted cancer gene probability $\hat{y}=0.99,0.98$ respectively) and two newly predicted cancer genes (COL5A1 and MSLN, predicted cancer gene probability $\hat{y} = 0.98,0.90$ respectively; see Figure \ref{fig3}A).
Notably, different genes were predicted as cancer genes leveraging evidence from different PPI networks.
For example, Multinet contributed to BRCA1, but not for MSLN. The newly predicted cancer gene COL5A1 combined information from all six PPI networks.
To systematically examine if some PPI networks were statistically more important contributors than others, we performed an ANOVA test across all meta edge importance for cancer genes (Figure \ref{fig3}B).
We found that different PPI networks made significantly different contributions to cancer gene prediction (P-value=$1.2e-65$). This suggests that certain PPI networks were more informative than others, likely due to their unique connectivity patterns that were more reflective of cancer development and progression.
Among pairwise comparisons, the updated version of Iref(2015) achieved a significantly higher contribution than the original Iref (P-value=$1.3e-50$, t-test), which consolidated our observation that the incorporation of other PPI networks substantially improved upon the model using only Iref (see Iref performance over input PPI networks in Figure \ref{fig2}). 
Thus, EMGNN successfully learned complementary information from the connectivity patterns in each layer of the multilayer graph, as shown by the overall contributions from individual graphs and gene-specific variations.

We hypothesized that the variation of meta-edge importance was a result of different cancer gene neighborhood in different PPI networks (Figure \ref{fig3}C). 
For a target gene, we define "cancer neighbors" as the number of neighboring genes that are known cancer genes, which is then normalized by the degree of the target node in the given network. 
Hypothetically, for gene A whose neighbors were all cancer genes across PPI networks, the meta-edge importances were also comparable.
In contrast, for gene B whose cancer gene neighbors varied across PPI networks, we should observe a positive covariance between cancer gene neighbors and meta-edge importance. 
To test our hypothesis, we computed the standard deviations of meta-edge importance for each labeled cancer gene, as well as the Pearson correlation between meta-edge importance and the cancer gene neighbors across PPI networks. 
For genes with large standard deviation of meta-edge importance, the important PPI networks tend to have a higher cancer neighbors, as demonstrated by a higher correlation between meda-edge importance and cancer neighbors (Figure \ref{fig3}D). 
Due to the complex graph convolutions that enabled message-passing beyond one-hop neighbors, our cancer neighbors may not fully capture the variations of PPI network importance. 
Overall, our empirical analysis demonstrates that cancer neighborhood is implicated in genes with divergent connectivity patterns across PPI networks.

\begin{figure*}[t]
    \centering
    \includegraphics[width=\textwidth]{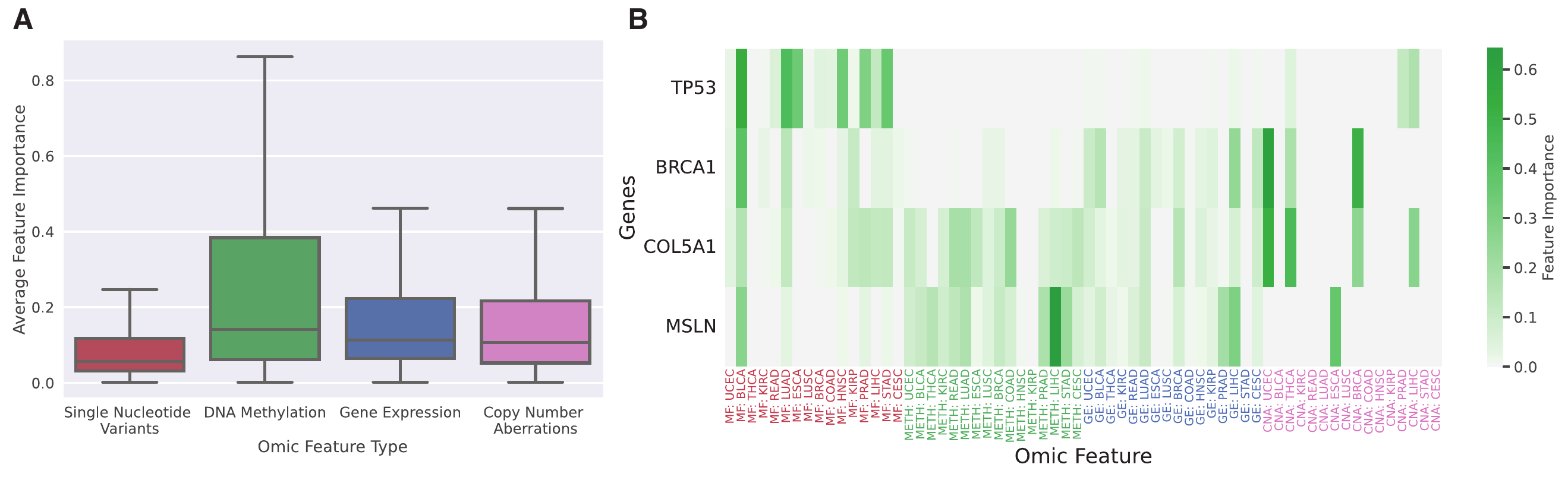}
    \caption{Explanations of multi-omic node features importance in cancer gene predictions.
    A) Overall distribution of node feature importance grouped by omic feature types, including single-nucleotide variants (MF), DNA methylation (METH), gene expression (GE) and copy number aberrations (CNA), for known cancer genes.
    B) Detailed node feature importance for the four genes analyzed in Figure \ref{fig3}B. X-axis labels were color-coded to match the omic feature types in panel A. Individual tumor types were coded according to TCGA study abbreviations \citep{weinstein2013cancer}. 
    }
\label{fig4}
\end{figure*}

On the individual gene level, a detailed explanation will reveal gene-specific genetic and molecular aberrations that the EMGNN model relies on for cancer gene prediction.
As EMGNN's node features were derived by multi-omics pan-cancer datasets, we next assessed if certain types of omic data were informative to cancer gene predictions. 
Our model explanation results of node features indicated that Single Nucleotide Variation (SNV) features were found to be significantly less informative than other types of features, which is consistent with previous reports that CNAs were more detrimental to cancer progression than SNVs \citep{elmarakeby2021biologically}.
In contrast, we observed that DNA methylation was significantly more important for known cancer gene prediction than other omics data (P-value<0.01 for all three pairwise t-test of other omics against DNA methylation).
We further examined the node feature importance of the same four genes (TP53, BRCA1, COL5A1, MSLN).
As expected, the omics feature contributions varied on different genes and were highly gene specific, demonstrating the heterogeneity of tumorigenesis.
Point mutations were major contributors to the prediction of TP53 as a cancer gene,  which is consistent with findings from previous studies \cite{almeida2009polymorphisms,guimaraes2002tp53}.
Moreover, the prediction of BRCA1 correctly identified gene expression and copy number aberrations as the most significant features~\cite{chang2022high}.  
DNA methylation had a moderate contribution for the two newly predicted cancer genes, COL5A1 and MSLN (Figure \ref{fig4}B).
As DNA methylations are reversible epigenetic modifications, this may suggest potential novel therapeutic targets for certain cancer genes mediated by DNA methylation \citep{sharma2010epigenetics}.

\subsection{EMGNN identifies newly predicted cancer genes by integrating multilayer graphs}

A key utility of EMGNN, given its superior performance and explainability, is to identify newly predicted cancer genes that share similar topological patterns to known cancer genes, but may have been missed by a conventional recurrent alteration analysis \citep{schulte2021integration, sherman2022genome}.
We applied the trained EMGNN model to predict cancer genes on a total of $n=14019$ unlabeled genes. 

By integrating multilayer graphs, EMGNN addressed the divergent and inconsistent cancer gene predictions from previous models trained using a single PPI network.
Prior to EMGNN, models trained using single PPI networks such as EMOGI had made conflicting predictions on which genes were cancerous.
Indeed, we observed substantial variations among the predictions of EMOGI models trained on individual PPI networks, with an average of $29\%$ and $63\%$ difference between the highest and lowest predictions for the top-100 and all unlabelled nodes, respectively. 
Furthermore, we found an average standard deviation of 25.2\% in unlabelled cancer gene predictions of EMOGI, demonstrating the predicted novel cancer genes were different when using different PPI networks. In contrast, EMGNN resolved these discrepancies using a more accurate and robust representation of the data from multilayer graphs.

Our analysis identified 435 genes with a high probability of being newly predicted cancer genes, with an 88\% predicted cancer gene probability threshold. This threshold was selected based on its ability to provide a precision greater than 95\% in the labeled set, indicating a high degree of accuracy in identifying true cancer genes. The identification of these novel cancer genes may provide new insights into the molecular mechanisms of cancer and offer potential targets for the development of novel therapies, demonstrating that machine learning predictions can augment the completeness of cancer gene catalogs \citep{schulte2021integration}. 
The complete list of gene predictions from EMGNN can be found in the project repository on \href{https://github.com/zhanglab-aim/EMGNN}{\color{blue}GitHub}.

\begin{figure*}[t]
    \centering
    \includegraphics[width=\textwidth]{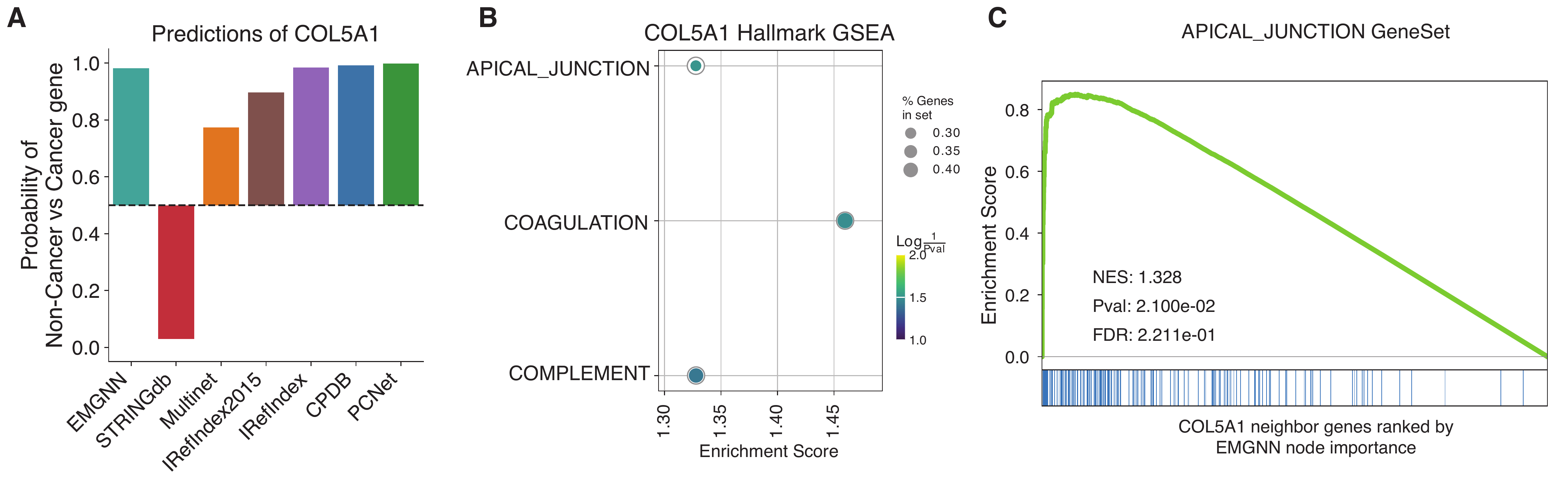}
    \caption{EMGNN predicts COL5A1 as a novel cancer gene and reveals biological insights.
    A) A comparison of predicted cancer gene probability from EMGNN and EMOGI models trained on single PPI networks. As a probability of 50\% equaled random guessing between cancer vs non-cancer gene, the bar heights reflected the prediction confidence.
    B) Three cancer hallmark genesets were significantly enriched in the important neighboring genes of COL5A1 as revealed by interpreting EMGNN model. 
    C) Enrichment of apical junction cancer hallmark geneset in COL5A1 neighboring genes. The neighboring genes of COL5A1 were ranked by their EMGNN node importance on the x-axis, where each blue bar represented a gene in the apical junction geneset. A strong left-shifted curve demonstrates enrichment of apical junction geneset in the top important genes to predict COL5A1 as a cancer gene.
    }
\label{fig5}
\end{figure*}

As a case study, we analyzed the predictions of a newly predicted cancer gene, COL5A1 (Figure \ref{fig5}). For this gene, EMOGI model trained on STRINGdb predicted a non-cancer gene with high confidence ($\hat{y}=0.03$), EMOGI models trained on IRefIndex, CPDB and PCNet predicted a cancer gene with high confidence ($\hat{y}>0.98$), while the models trained on Multinet and IRefIndex2015 predicted a cancer gene with moderate likelihood ($\hat{y}=0.775$ and $0.897$, respectively).
The fact that STRINGdb was the best-performing PPI network among models trained on individual PPI networks further complicated the decision making whether COL5A1 should be considered as cancer/non-cancer gene (Table \ref{table1}).
This level of divergence in predictions hinders a trustworthy adaptation of model predictions in clinical and pragmatic settings.
In contrast, EMGNN integrated the information from each individual PPI network in a data-driven approach and provided more accurate, unified predictions of cancer genes (Table \ref{table1}). EMGNN predicted COL5A1 as a cancer gene with high confidence (Figure \ref{fig5}A). Importantly, we also found all individual PPI networks were contributing similarly to the final EMGNN prediction (Figure \ref{fig3}A), and revealed the multi-omics features implicated in this prediction (Figure \ref{fig4}B). 

Leveraging the explanation results of node contributions for COL5A1 across its neighboring genes, we further illustrated the potential biological mechanisms of COL5A1 using a gene set enrichment analysis (Methods). 
These neighboring genes formed a ranked gene list based on their explained EMGNN contributions to the prediction of COL5A1 as a cancer gene or not. 
We discovered that three cancer hallmark gene sets, i.e. apical junction, coagulation, and complement system (part of the innate immune system), were significantly enriched in COL5A1 neighboring genes (Figure \ref{fig5}B). 
For example, the apical junction cancer hallmark geneset contained genes annotated to function in cell-cell adhesion among epithelial cells, many of those were enriched in the top contributors (Figure \ref{fig5}C).
This was further supported by other studies, where COL5AI was associated with skin cancer, the type of cancer with a strong epithelial cell origin \citep{mann2015transposon,zhu2022hypoxia,gu2021col5a1}.
Therefore, we demonstrated how molecular mechanisms of newly predicted cancer genes could be interpreted and discovered using the explainable EMGNN framework.

\section{Discussion}

The biomedical and biological domain contains a wealth of information, which is often represented and analyzed using graph structures to reveal relationships and patterns in complex data sets.
Various gene interaction and protein-protein interaction networks describe the functional relationships of genes and proteins, respectively.
The gene-gene relationships were often described in generic cellular contexts and/or by integrating different, heterogeneous sources of information.
Therefore, a single graph often struggles to best match disease-specific conditions, and different graph construction and integration methods render distinct predictive powers~\citep{cao2014new,cho2016compact}.
Substantial efforts have been devoted to develop integrated~\citep{cao2014new,cho2016compact} and tissue-specific graphs ~\citep{zitnik2017predicting}.
Here, we took a complementary approach and developed a new graph learning framework, EMGNN, to jointly model multilayered graphs.
Applying EMGNN to predict cancer genes demonstrated its superior performance over previous graph neural networks trained on single graphs.
We also employed model explanation techniques to assess both node and edge feature importance.
Our results showed that EMGNN leveraged the complementary information from different graph layers and omics features to predict cancer genes.
Importantly, we found that cancer genes that have conflicting predictions based on different single graphs, or are missed by  previous state-of-the-art predictors, can be recovered effectively using EMGNN.
This demonstrates the robustness of EMGNN predictions by joint modeling the multilayered graphs.

The EMGNN model can be viewed as a data-driven, gradient-enabled integration method for multiple graphs.
By providing multiple PPI networks as input, EMGNN learns from the different connectivity patterns that represent complementary information to predict cancer genes.
Since all PPI networks share the same type of nodes and edges (though not necessarily the same set of nodes in each network), EMGNN currently integrates homogeneous, undirected graphs; however, the EMGNN framework can be extended to various types of graphs and to perform cross-data modality integration.
In biology and biomedicine, hierarchical graphs and heterogeneous graphs are particularly prevalent, such as Gene Ontology\citep{gene2004gene}.
For example, biomedical data is often organized in hierarchical levels, starting with genes and molecules, moving on to cells and tissues, and finally reaching the level of individual patients and populations.
Therefore, an interesting future direction is to apply EMGNN to model multiple graphs with more heterogeneous node and edge types, and with more complex inter-graph structures.

DNA methylation and gene expression aberrations are major contributors to EMGNN's cancer gene predictions, which are considered important features when explaining its omics node features.
Unlike single nucleotide variation and copy number alteration that introduced permanent mutations to DNAs, epigenetic and transcriptomic alterations of cancer genes are potentially reversible by targeted therapies.
Model explanations for EMGNN revealed these molecular aberrations that may be leveraged for screening and re-purposing of drugs, especially for previously less well-characterized, newly predicted cancer genes.
This highlights the importance of model explanations to gain biological and biomedical insights when developing deep learning models to predict gene pathogenicity.

In summary, we present a novel deep learning approach for the prediction of cancer genes by integrating multiple gene-gene interaction networks. 
By applying graph neural networks to each individual network and then combining the representations of the same genes across networks through a meta-graph, our model is able to effectively integrate information from multiple sources. 
We demonstrate the effectiveness of our approach through experiments on benchmark datasets, achieving state-of-the-art performance.
Furthermore, the ability to interpret the model's decision-making process through the use of integrated gradients allows for a better understanding of the contribution of different multi-omic features and PPI networks. 
Overall, our approach presents a promising avenue for the prediction of novel cancer genes.



\section{Methods}
\subsection{Datasets}
We use the datasets and train/test splits compiled by Schulte-Sasse et al. \citep{schulte2021integration} to ensure a fair comparison.
Specifically, we trained our proposed model with six PPI Networks: CPDB~\citep{kamburov2011consensuspathdb}, Multinet~\citep{khurana2013interpretation}, PCNet~\citep{huang2018systematic}, STRING-db~\citep{szklarczyk2019string}, Iref~\citep{razick2008irefindex} and its newest version Iref(2015). 
The preprocessing of these six PPI networks was done in Schulte-Sasse et al. \citep{schulte2021integration}. We provide a brief description of the preprocessing steps here for clarity and self-containment.
Depending on the source of the data, different confidence thresholds were applied to filter out low-confidence interactions. 
Interactions with a score higher than 0.5 in CPDB and 0.85 in STRING-db were included. For Multinet and IRefIndex(2015), the data was obtained from the Hotnet2 github repository~\citep{reyna2018hierarchical}. 
In the case of the recent version of IRefIndex, analysis was restricted to binary interactions between two human proteins. No further processing was performed on the PCNet.

As node features, we used single-nucleotide variants (MF), copy number aberrations (CNA), DNA methylation (METH) and gene expression (GE) data of $29,446$ samples from TCGA~\citep{weinstein2013cancer}, from $16$ different cancer types.  
The SNV frequency was calculated for each gene in each cancer type by dividing the number of non-silent SNVs in that gene by its exonic gene length. 
Gene-associated CNAs were collected from TCGA, where the copy number rate for each gene was defined as the number of times it was amplified or deleted in a specific cohort. 
DNA methylation for each gene in a cancer type was computed as the difference in methylation signal between tumor and matched normal samples.
The expression level of each gene in each sample was quantified using RNA-seq data~\citep{wang2018unifying}, and differential expression was computed as a $\log_2$ fold change between cancer and matched normal samples, averaged across samples.

The positive examples included expert-curated cancer genes, high-confidence cancer genes mined from PubMed abstracts, and genes with altered expression and promoter methylation in at least one cancer type~\citep{repana2019network,sondka2018cosmic}. The negative examples were compiled by removing genes not associated with cancer from a set of all genes and were selected based on their absence in the positive set and various cancer databases~\citep{repana2019network,mckusick2007mendelian}. 
For further information regarding the data collection and processing methods, refer to the study by Schulte-Sasse et al. \cite{schulte2021integration}.

\subsection{Multilayer Graph Neural Network}
\noindent \textbf{GNNs.} Let a graph be denoted by $G = (V,E),$ where $V = \{v_1,\ldots,v_N\}$ is the set of vertices and $E$ is the set of edges. Let $\mA \in \mathbb{R}^{N \times N}$ denote the adjacency matrix, $\mX=[x_1,\ldots,x_N]^T \in \mathbb{R}^{N \times d_I}$ denote the node features where $d_I$ is the features dimensions and $\bm{Y} = [y_1,\ldots,y_N]^T \in \mathbb{N}^{N}$ denote the label vector. 
Graph neural networks have been successfully applied to many graph-structured problems\citep{4700287,fung2021benchmarking}, as they can effectively leverage both the network structure and node features. They typically employ a message-passing scheme, which constitutes of the two following steps. In the first step, every node aggregates the representations of its neighbors using a permutation-invariant function. In the second step, each node updates its own representation by combining the aggregated message from the neighbors with its own previous representation,
\begin{align}
    m_u^{(\layersuperscript)} &= \text{Aggregate}^{(\layersuperscript)} \left( \left\lbrace \mathbf{h}_v^{(\layersuperscript-1)} : v \in \neighborhood{u} \right\rbrace \right),\numberthis \label{eqn:aggregation}\\
    \mathbf{h}_u^{(\layersuperscript)} &= \text{Combine}^{(\layersuperscript)} \left( \mathbf{h}_u^{(\layersuperscript-1)}, \mathbf{m}_u^{(\layersuperscript)} \right),
\end{align}
where $h_u^{(\layersuperscript)}$ represents the hidden representation of node $u$ at the $\layersuperscript^{\mathrm{th}}$ layer of the GNN architecture. Many choices for the \textit{Aggregate} and \textit{Combine} functions have been proposed in the recent years, as they have a huge impact in the representation power of the model\citep{xu2018how}. Among the most popular architectures, are Graph Convolution Networks (GCNs)~\citep{kipf2017semisupervised}, 
and Graph Attention Networks (GAT)~\citep{velikovi2017graph}. In GCN, each node aggregates the feature vectors of its neighbors with fixed weights inversely proportional to the central and neighbors’ node degrees,$
\mathbf{h}^{\prime}_u = \mathbf{\mW}^{\top} \sum_{v \in
\mathcal{N}(u) \cup \{ u \}} \frac{\mathbf{h}_v}{\sqrt{\hat{d}_v
\hat{d}_u}}$,
with $\hat{d}_i = 1 + \sum_{j \in \mathcal{N}(i)} 1$.
In GAT, each node aggregates the messages from its neighbor using learnable weighted scores: 
$ \mathbf{h}^{\prime}_u = \alpha_{u,u}\mathbf{\mW}\mathbf{h}_{u} + \sum_{v \in \mathcal{N}(u)} \alpha_{u,v}\mathbf{\mW}\mathbf{h}_{v}
$, where the attention coefficients $\alpha_{u,v}$ are computed as \\ $\alpha_{u,v} =
\frac{
\exp\left(\mathrm{LeakyReLU}\left(\mathbf{a}^{\top}
[\mathbf{\mW}\mathbf{h}_u \, \Vert \, \mathbf{\mW}\mathbf{h}_v]
\right)\right)}
{\sum_{k \in \mathcal{N}(u) \cup \{ u \}}
\exp\left(\mathrm{LeakyReLU}\left(\mathbf{a}^{\top}
[\mathbf{\mW}\mathbf{h}_u \, \Vert \, \mathbf{\mW}\mathbf{h}_k]
\right)\right)}.$

\noindent \textbf{Multilayer Graph Construction.}
Extending graph neural networks to handle multiple networks is not a trivial task, as they are designed to operate on a single graph. 
Next, we describe our method, which can accurately learn node representations using graph neural networks, from multilayer graphs.

Let $N$ be the total number of genes, each associated with a feature vector $x_j \in R^{d_I}$. Let also $K$ be the number of gene-gene interaction networks. We represent each network $G^{(i)}$ with an adjacency matrix $A^{(i)} \in \bm{Z}^{N_i \times N_i}$ and feature matrix $X^{(i)} \in R^{N_i \times d_I}$, where $N_i$ is the number of genes in the $i$-th network. Since some genes are not presented in all the graphs, the following equation holds $N_i \leq N$, $i\in \setk{K-1}$.

In the first step, for each graph $G^{i}$ we apply a graph neural network $f_1$ that performs message-passing and updates the node representation matrix $H^{(i)} = f_1(X^{(i)}, A^{(i)})$ of each graph $i\in \setk{K-1}$. 
We set $f_1$ to be shared across all graphs. This design allows us to handle a variable number of graphs while keeping the number of trainable parameters fixed.

Next, to aggregate and share information between each graph, we construct a meta graph $G_{meta, j}$ for each gene/node $j$, where the same genes $j$ across all graphs are connected to a meta node $v_j$.
We initialize the features of the meta node $v_j$ with the initial features of the corresponding gene $j$.
We apply a second GNN $f_2$ to update the representation of the meta node $v_j$, $H_{meta,j} = f_2(X_{meta,j}, A_{meta,j})$, where $X_{meta,j}$ contains the features of gene $j$ from all the networks and $A_{meta,j}$ is the adjacency matrix of the meta graph $G_{meta, j}$, $j\in \setk{N}$. We set $f_2$ to be shared across all genes. Therefore, in this stage, the model combines and exchanges information between the different networks. Finally, a multi-layer perceptron $f_3$ predicts the class of the meta node $j$, $\hat{y_j} =f_3(H_{meta,j})$. An illustration of the proposed model can be found in Figure \ref{fig1}.

\noindent \textbf{Experimental Details.}
To ensure a fair comparison with previous work, we utilized the same experimental setup as in Schulte et al. \citep{schulte2021integration}. In particular, we divided the data for each testing graph into a 75\% training set and a 25\% testing set using stratified sampling to maintain equal proportions of known cancer and non-cancer genes in both sets. Since our model takes multiple graphs as input for each experiment, we retained the test nodes of one graph as the test set, and we allocated 90\% of the remaining nodes from the other graphs to the training set and 10\% to the validation set. 
When adding other PPI networks to the training and validation set, we held-out the same testing set from the combined training set and kept the test set identical to previous works \citep{hong2022reusability, schulte2021integration}.
An illustration of the process of adding a new biological network and a definition of training and validation splits are shown in Supplementary Figure 1. 
The model was trained for $2000$ epochs, using the cross-entropy loss function, and the ADAM optimizer~\citep{kingma2014adam} with a learning rate of $0.001$. The initial GNN had three layers with a hidden dimension of $64$, while the meta-GNN had a single layer with a hidden dimension of $64$. 
The Supplementary Table 1 
provides information on the PPI networks, including statistics, as well as the number of positive and negative genes used for training. 


\subsection{Model interpretation}
We used the integrated gradient (IG) module in Captum, to assign an importance score to each input feature.
Captum is a tool for understanding and interpreting the decision-making process of machine learning models \citep{kokhlikyan2020captum}. It offers a range of interpretability methods that allow users to analyze the predictions made by their models and understand how different input variables contribute to these predictions. 
IG interprets the decisions of neural networks by estimating the contribution of each input feature to the final prediction. The integrated gradient approximates the integral of gradients of the model's output with respect to the inputs along a straight line path from a specific baseline input to the current input. The baseline input is typically chosen to be a neutral or a meaningless input, such as an all-zero vector or a random noise.
Formally, let $F(x)$ be the function of a neural network, where $x$ is the input and $\hat{x}$ is the baseline input. 
The integrated gradients for input $x$ and baseline $x0$ along the $i$-th dimension is defined as:
  $\texttt{IntegratedGrads}_i(x) = (x_i-\hat{x_i})\int_{\alpha =0}^1 \frac{\partial F(\hat{x}+\alpha (x-\hat{x})}{\partial x_i} d\alpha$,
where the integral is taken along the straight line path from $\hat{x}$ to $x$ and $\partial F(x) / \partial x_i$ is the gradient of $F(x)$ along the $i$-th dimension.
    
However, the traditional integrated gradient method, which is designed for single input models, is not directly applicable to graph neural networks as they have two distinct inputs, namely node features and network connectivity. This necessitates the development of a modified approach for computing integrated gradients in graph neural networks that considers both inputs. To this end, we propose a decomposition of the problem into two parts: identifying the most important node features and identifying the most crucial edges in the network separately. 
Since we predict the class of each gene  by combining all the graphs, from the meta-node representations, we apply the interpretation analysis only to the meta-nodes.

\noindent \textbf{Node feature interpretation analysis.}
We analyze the contribution of node features to the predictions of the GNN by using the traditional integrated gradient method while keeping the edges in the network fixed. 
Specifically, we interpolate between the current node features input and a baseline input where the node features are zero: $ Attribution_{x_i} = (x_i-\hat{x_i})\int_{\alpha =0}^1 \frac{\partial F(\hat{x}+\alpha (x-\hat{x},A)}{\partial x_i} d\alpha$, where $A$ are the adjacency matrices of the graphs.
Since the prediction for each gene is also based on the features of surrounding genes in the graphs, we extract attribution values for the $k$-hop neighbor genes as well, where $k$ is equal to the number of message-passing layers in the first GNN. 
Therefore, the output of the attribution method for each node $u$, is a matrix $\mathbf{K^{(u)}} \in \mathbb{R}^{N\times d}$. Each entry $K_{ij}$ of the matrix, corresponds to the attribution of the feature $j$ of node $i$ to the target node $u$. 
From this matrix, we select the row that corresponds to the feature attributions of the corresponding meta node.

\noindent \textbf{Edge feature interpretation analysis.}
To analyze the contribution of edges in the meta-graph to the predictions of the GNN, we use the integrated gradient method for the edges while keeping the node features fixed. Specifically, we interpolate between the current edge input and a baseline input where the weights of the edges are zero: 
$ Attribution_{e_i} = \int_{\alpha =0}^1 \frac{\partial F(X,A_\alpha)}{\partial w_{e_i}} d\alpha$, where $A_\alpha$ corresponds to the graphs with the edge weights equal to $\alpha$.
We further normalize the attribution values of each meta node by dividing them by their maximum value, resulting in a range of [0, 1] for each edge.
This explanation technique allows us to understand which edges in the meta-graph are crucial for the model's decision-making process, and therefore which input PPI networks are important for each gene prediction.

\subsection{Newly predicted cancer gene discovery}
We applied the trained EMGNN model that combined all six individual PPI networks to predict novel cancer genes in the $n=14019$ unlabeled genes.
We ranked these genes by their predicted cancer gene probability for potential new predicted cancer genes (NPCG) in this study.
For each unlabeled gene, we also applied EMOGI models trained on individual PPI networks to predict the probability of it being a cancer gene. The complete list of the predicted probabilities for all the unlabeled genes, can be found in the project repository on \href{https://github.com/zhanglab-aim/EMGNN}{\color{blue}GitHub}.


\subsection{Gene set enrichment analysis}
To understand the biological mechanisms of EMGNN's cancer gene prediction, we employed gene set enrichment analysis (GSEA) to analyze the functional enrichment of important gene features in curated cancer pathway annotations. 
Specifically, to determine the importance of neighboring gene nodes, we aggregated the maximum feature importance of each node using Captum's feature explanation results.
Genes with zero importance were excluded in this analysis as they did not contribute to the prediction of this target gene.
We then ranked the neighboring gene nodes based on their importance and used this ranked gene list as input for GSEA.
The enrichment p-value and multiple testing corrected FDR were computed by GSEA python package \citep{fang2023gseapy} against cancer hallmark gene sets \citep{liberzon2015molecular}.


\section*{Acknowledgements}
We thank all members of the Zhang laboratory for helpful discussions. This work was performed at the high-performance computing resources at Cedars-Sinai Medical Center and the Simons Foundation.

\section*{Funding}
This work has been supported by an institutional commitment fund from Cedars-Sinai Medical Center to ZZ. 

\bibliographystyle{plain}
\bibliography{main}

\end{document}